\documentclass{article}

\usepackage{xspace}
\usepackage{enumitem}
\usepackage{mathtools}
\usepackage{mdframed}
\usepackage{xcolor}
\usepackage{wrapfig}
\usepackage[english]{babel}
\usepackage{amssymb}
\usepackage{tikz}
\usetikzlibrary{shapes}
\usepackage{pgfplots} %
\usepackage[margin=2.8cm]{geometry}
\usepackage[colorlinks = true]{hyperref} 

\newcommand\blfootnote[1]{%
	\begingroup
	\renewcommand\thefootnote{}\footnote{#1}%
	\addtocounter{footnote}{-1}%
	\endgroup
}

\newcommand{\bit}{\ensuremath{\{ 0, 1 \}}}
\newcommand{\ie}{\textit{i.e.}\@\xspace}
\newcommand{\eg}{\textit{e.g.}\@\xspace}
\newcommand{\R}{\ensuremath{\mathbb{R}}}
\newcommand{\Z}{\ensuremath{\mathbb{Z}}}

\newcommand{\M}{\ensuremath{\mathcal{M}}}
\newcommand{\I}{\ensuremath{\mathcal{I}}}
\newcommand{\C}{\ensuremath{\mathcal{C}}}

\newcommand{\tr}{\top}

\newcommand{\sign}{\textup{sign}}
\renewcommand{\vec}[1]{\mathbf{#1}}
\newcommand{\sk}{\vec{sk}}
\newcommand{\pk}{\vec{pk}}
\newcommand{\enc}{\mathsf{Enc}}
\newcommand{\dec}{\mathsf{Dec}}
\newcommand{\gen}{\mathsf{Gen}}
\newcommand{\argmax}{arg\,max}
\newcommand{\mult}{\operatorname{\mathsf{Mult}}}
\newcommand{\MP}{\textup{P}}

\newcommand{\cli}{\textup{C}}
\newcommand{\Pte}{\Pi_{\textup{TE}}}
\newcommand{\Ppps}{\Pi_{\textup{PPS}}}
\newcommand{\Psc}{\Pi_{\textup{SC}}}
\newcommand{\Tenc}[1]{[\,#1\,]}

\newcommand{\qquote}[1]{“#1”}

\begin{document}

\title{Privacy-Preserving Collaborative Prediction using Random Forests}

\author{Irene Giacomelli$^1$
	\and 
	Somesh Jha$^2$
	\and 
	Ross Kleiman$^2$
	\and 
	David Page$^2$ 
	\and 
	Kyonghwan Yoon$^2$\\[.1cm]
$^1$ISI Foundation, Turin, Italy; $^2$University of Wisconsin-Madison, Madison, WI, US}

\date{\today}

\maketitle              

\pagestyle{plain}

\blfootnote{This is the full version of the paper presented at the AMIA Informatics Summit 2019. Large part of this work was done while IG was a research assistant at the University of Wisconsin-Madison.}

\begin{abstract}
	
	We study the problem of privacy-preserving machine learning (PPML) for ensemble methods, focusing our effort on random forests. In collaborative analysis, PPML attempts to solve the conflict between the need for data sharing and privacy. This is especially important in privacy sensitive applications such as learning predictive models for clinical decision support from EHR data from different clinics, where each clinic has a responsibility for its patients’ privacy. 
	We propose a new approach for ensemble methods: each entity learns a model, from its own data, and then when a client asks the prediction for a new private instance, the answers from all the locally trained models are used to compute the prediction in such a way that no extra information is revealed. We implement this approach for random forests and we demonstrate its high efficiency and potential accuracy benefit via experiments on real-world datasets, including actual EHR data.
	
\end{abstract}

\section{Introduction}
\label{sec:intro}

Nowadays, machine learning (ML) models are deployed for prediction in many privacy sensitive scenarios (\eg, personalized medicine or genome-based prediction).  A classic example is disease diagnosis, where a model predicts the risk of a disease for a patient by simply looking at his/her health records.
Such models are constructed by applying learning methods from the literature to specific  data collected for this task (the training data,---instances for which the outcome is known---in the preceding example these are health records of patients monitored for the specific disease). Prior experience in ML model training suggests that having access to a large and diverse training dataset is a key ingredient in order to enhance the efficacy of the learned model (\eg, see \cite{warf09}). A training dataset with these feature can be  created by merging several silos of data collected locally by different entities. Therefore, sharing and merging data can result in mutual gain to the entities involved in the process and, finally, to the broader community. For example, hospitals and clinics located in different cities across a country can locally collect clinical data that is then used to run a collaborative  analysis with the potential to improve the health-care system of the entire country.  However, in privacy sensitive scenarios, sharing data is hindered by significant privacy concerns and legal regulations (\eg, HIPAA laws in the United~States and  GDPR for the European Union). In the example described before, sharing clinical data directly competes with the need for healthcare providers to protect the privacy of each patient and respect current privacy policies and laws.	

Based on the preceding discussion, we often face the following dilemma: share data to improve accuracy or keep data and information secret to protect privacy?
Notice that de-identification cannot resolve this standoff: several	works  demonstrated that sharing de-identified data  is not a secure	approach since in many contexts the potential for re-identification is high (\eg, \cite{homer2008resolving,NS08}). More sophisticated anonymization criteria (\eg, $k$-anonimity, $l$-diversity, $t$-closeness, etc.) were proposed by the database community. While arguably better than de-identification, all such “syntactic” approaches work only in presence of assumptions regarding the adversary’s background knowledge. Conversely, cryptographic tools  can guarantee perfect privacy of shared data in more general situations. 
For example,  a number of \emph{privacy-preserving training} algorithms have been proposed since the seminal paper of Lindell and Pinkas \cite{LP00} introduced this concept in 2000. 
Existing proposals use different cryptographic tools (\eg,  homomorphic encryption and multi-party computation) in order to allow different parties to run known learning algorithms on the merge of local datasets without revealing the actual data. This approach guarantees privacy for all data-providers involved at the price of high communication and computation overhead. 
Once the model is learned, we face another privacy problem: using the model to compute a prediction for inputs while both the model and the input data are sensitive information privately held  by  different parties. This problem can be solved using again cryptographic tools, and an algorithm designed for this task is called \emph{privacy-preserving scoring}. This is a two-party protocol where a provider with a  proprietary model interacts with a client with a private input in order to score the model on the input without revealing neither the model nor the input.
In conclusion, a solution that uses the current tools to guarantee privacy at all levels (\eg, for the data providers, model providers, model users)  deploys  two privacy-preserving systems, a first one for training and a second one for scoring.

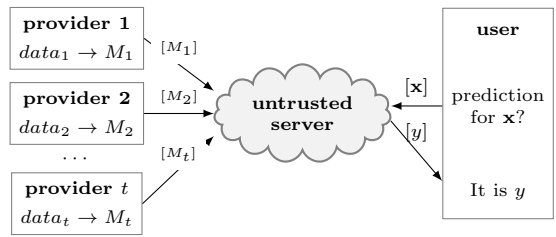
\begin{wrapfigure}{R}{0.5\textwidth} 
	\centering
	\begin{tikzpicture}[font=\scriptsize]
	\node[draw=gray, align=center](h1){\textbf{provider 1}\\[.1cm]
		$data_1\rightarrow M_1$};
	\node[draw=gray,  below of = h1, align=center](h2){\textbf{provider 2}\\[.1cm]
		$data_2\rightarrow M_2$};
	\node[below of = h2, node distance=.6cm](h){$\cdots$};
	\node[draw=gray, below of = h2, node distance=1.2cm, align=center](hn){\textbf{provider $t$}\\[.1cm]
		$data_t\rightarrow M_t$};
	
	\node [right of = h2, node distance=3cm, fill=gray!10, cloud, cloud puffs=15, cloud ignores aspect, minimum width=2cm, minimum height=1cm, draw=gray, thick, align=center] (server1) {\textbf{untrusted} \\\textbf{server}};
	
	
	\node[draw=gray, right of=server1, node distance=2.6cm, minimum height=2.8cm, align=center](user){\textbf{user}\\[.6cm]prediction\\ for $\vec{x}$?\\[.7cm]It is $y$};
	
	\draw[-latex]([yshift=.1cm]user.west) -- node[above] {$[\vec{x}]$} ([yshift=.1cm]server1.east);		
	\draw[-latex]([yshift=-.1cm]server1.east) -- node[above] {$[y]$} ([yshift=-.9cm]user.west);
	
	\draw[-latex](h1.east) -- node[above,fill=white, font=\tiny] {$[M_1]$} ([yshift=.3cm]server1.west);		
	\draw[-latex](h2.east) -- node[above, font=\tiny] {$[M_2]$} (server1.west);		
	\draw[-latex](hn.east) -- node[above, fill=white, font=\tiny] {$[M_t]$} ([yshift=-.3cm]server1.west);	
	;	
	\end{tikzpicture}
	\caption{Overview of the new \qquote{locally learn then merge} approach in the cloud model. The providers upload the encrypted models to the server and then go off-line. The server is on-line to answer to the prediction requests of the user.
	}
	\label{fig:1}
\end{wrapfigure}

%
%
%
%
%
%

In this work, we notice that for \emph{ensemble methods}, for which the learned model is formed by a set of more basic models and the prediction for a new instance is computed by blending together the basic predictions,  there can be an easier and more efficient solution that needs only one system; we refer to this solution as the \emph{\qquote{locally learn then merge}} approach. Each entity with a local data silo (\ie, providers) can train its own local model $M_i$, and then the prediction given by these models can be merged at the moment when the scoring for a new (eventually private) instance is computed. That is, a user with input $\vec{x}$ gets $y=\Phi(M_1(\vec{x}),\dots, M_t(\vec{x}))$ for a specific merging function $\Phi$. Here $M_i(\vec{x})$ indicates the prediction of the local model $M_i$ for the instance $\vec{x}$.
In this approach, privacy concerns coming from data sharing in the training phase are not present since, clearly, local training does not require data sharing. Moreover, there is no overhead for the training phase (this is run as in the standard ML scenario), while the final prediction can benefit from merging the local predictions via the function $\Phi$.    
On the other hand, accuracy loss (with respect to a model learned from the merged data) and information leakage can happen during the merging/scoring phase. 
In particular, a challenge remains with this simple and elegant approach to collaborative ML: if we want to guarantee model and user's input privacy (\ie, the user learns $y$ and no other information on the models $M_i$, the providers learn nothing about $\vec{x}$), then even after the training phase each provider must maintain its own on-line server and communicate with the client and the other providers\footnote{
	Notice that  using $t$ instances of a given privacy-preserving scoring algorithm (once for each provider interacting with the user only) reveals $M_i(\vec{x})$ to the user. In this work, we want the user to know only the final prediction $y$ and no other  information on the local models.}
each time a new prediction is requested.
Since in a real-world scenario (\ie, healthcare environment), this requirement can be cumbersome to implement, we design our system in the \emph{cloud model}~\cite{BCF17}, where the computation of the prediction from the local models is outsourced to a central server and providers are not required to be on-line during the scoring process (Fig.\ref{fig:1}). 
Since we do not require the server to be trusted, each model $M_i$ is sent to the server in encrypted form (\ie, $[M_i]$).  Once this is done, the providers (\eg, clinics) can go off-line and when a user (\eg, medical research institution) requires access to the models to compute  predictions for  new data, the server  communicates with it and computes the answer from the encrypted models.

In this work, we specify and evaluate the \qquote{locally learn then merge} paradigm in the cloud model for a widely-used ensemble method: random forests.
\emph{Random forests} \cite{Breiman2001} are among the most accurate and widely-used ML ensemble models and are employed across a variety of challenging tasks, including predictive modeling from clinical data (\eg, \cite{lantz2016machine}), that are characterized by high dimension and variable interactions, or other non-linearities in the target concept.
A random forest is a collection of simple decision trees. By the use of different trees, a random forest can capture variable interactions without the need for the learner to know or guess all relevant interactions ahead of time in order to represent them with new variables (interaction terms); by their ensemble nature, random forests effectively reduce the over-fitting  often observed with ordinary decision tree learning.
A less-recognized advantage of random forests is that they can be learned in a distributed manner.
In this circumstance, separate random forests can easily be learned locally by entities with data silos, and then the prediction for a new instance is computed as the arithmetic mean of the predictions of all the trees in the locally trained random forests (\ie, the merging function $\Phi$ is the arithmetic mean). We design a system implementing  this approach for random forest using  standard and fast cryptographic primitives (linearly homomorphic encryption and oblivious transfer). As a special case our system  also addresses the previously-studied task of privacy-preserving scoring for a single forest.
While our scheme is efficient even for forests of many trees, not surprisingly its run-time and communication complexity grow exponentially with maximum tree depth in a forest.
Therefore we also provide empirical evidence that across a variety of data sets and tasks, increasing the number of trees can effectively make up for any accuracy or AUC lost by incorporating a stringent limit on tree depth, such as 8 or even~6.

\paragraph{Our Contribution and Paper Outline.}
In summary, our contributions are as follows:
\begin{itemize}
	\item We suggest that for ensemble models, where the output is naturally computed by blending together basic predictions, our new approach  \qquote{locally learn then merge}  
	can represent an effective alternative to the standard approach for privacy-preserving collaborative  ML. 
	\item In Section~\ref{sec:system}, we instantiate the new approach in the cloud model for random forests. In particular, we design ad-hoc procedures for the model encryption (off-line) and the decision-tree evaluation (on-line).
	A simplified version of our system can be seen as a standard privacy-preserving scoring algorithm for random forests or decision trees (see Appendix~\ref{app:pps}).
	\item Since the  efficiency of our system is influenced mainly by the depth of the decision trees learned,  in  Section~\ref{sec:hyper} we present empirical evidence that small depth random forest can achieve high efficiency.
	\item We empirically evaluate the efficacy of the proposed approach  and the efficiency of its implementation running experiments both on  UCI datasets and on real Electronic Health Record (EHR) data and using different computing resources. Section~\ref{sec:performance} presents the relative results.
\end{itemize}
The next section (Section~\ref{sec:background}) briefly recalls standard notions for decision trees and random forests. It also describes the notation and the cryptographic tools we use in this work.

\section{Background}
\label{sec:background}

\begin{wrapfigure}{R}{0.36\textwidth} 	
	\centering
	\begin{tikzpicture}[level distance=1cm,
	level 1/.style={sibling distance=2cm},
	level 2/.style={sibling distance=1cm}, scale=0.8]
	\tikzstyle{every node}=[circle,draw, scale=0.7]
	\node (Root)  {$N_1$}
	child{
		node[fill=orange!40] {$N_2$}  
		child { node[fill=orange!70] {$\ell_1$}}
		child { node[fill=orange!80!red] {$\ell_2$}}}
	child{
		node[fill=green!20] {$N_3$}
		child { node[fill=green!60] {$\ell_3$}}
		child { node[fill=green!80!blue] {$\ell_4$}}
	};	
	\end{tikzpicture}\\[.2cm]
	$\begin{aligned}
	P_{2,1}(x_1,x_2)=(x_1-1)(x_2-1)\\
	P_{2,2}(x_1,x_2)=(x_1-1)(x_2+1)\\
	P_{2,3}(x_1,x_3)=(x_1+1)(x_3-1)\\
	P_{2,4}(x_1,x_3)=(x_1+1)(x_3+1)
	\end{aligned}$
	\caption{Polynomial representation of  a complete binary tree of depth 2.}
	\label{ex1}
\end{wrapfigure}

\subsection{Decision Trees and Random Forests} 
\label{sec:dt}
Decision trees are a nonparametric ML model used for both classification and regression problems.
While there are a myriad of algorithms for constructing decision trees, we focus here on describing the model representation of the scoring procedure.
A decision tree, $T$, can be viewed as  mapping a column vector  $\vec{x}=(\vec{x}[1],\dots, \vec{x}[n])^\tr$ of features to a prediction value $y$.
In practice, we assume that $T$ is represented as a directed acyclic graph with two types of nodes: \emph{splitting nodes} which have children, and \emph{leaf nodes} which have no children. Moreover, $T$ has a single root node, which is also a splitting node, that has no parents.
For an input $\vec{x}\in \R^n$, we traverse the tree $T$ starting from the root and reach a leaf.
Each splitting node $N_i$ is defined by a pair $(j_i, t_i)$ where $j_i$ is an index in $\{1,\dots, n\}$ and $t_i\in\R$ is a threshold value. 
In the root-leaf path, at node $i$  we take the right branch if $\vec{x}[j_i]\geq t_i$. Otherwise, we take the left one.
Thus, each splitting node $N_i$ is associated with the function $N_i(\vec{x})=\vec{e}_{j_i}^\tr\cdot\vec{x}-t_i$ and the value $n_i=\sign(N_i(\vec{x}))$ (where $\cdot$ is the standard row-by-column multiplication).
Here the vector $\vec{e}_i$ is the column vector  in $\R^n$ with all zeros except for a 1 in position $i$ and  $\vec{e}_{i}^\tr$ is its transpose. Moreover, if $x\in\R$, then $\sign(x)=1$ if $x\geq 0$ and $\sign(x)=-1$ otherwise. 
In this way we traverse the tree and we reach  a leaf node.
The $i$-th leaf node is associated with the label $\ell_i$, which is defined to be the prediction of the query $\vec{x}$ that reaches the $i$-th leaf (\ie, $y=T(\vec{x})=\ell_i$). 
The  format of the labels $\{\ell_i\}_i$ depends on the specific ML problem (regression, multiclass classification or binary classification). 
In this work, we assume $\ell_i\in [0,1]$  representing the probability of $\vec{x}$ being classified as $+$ in a binary classification problems with labels $\{+,-\}$.  
We use the following \emph{indexing system} for a complete binary tree: The root node is the first splitting node, $N_1$, then we label the remaining splitting nodes level by level and in order from left to right. For the leaf nodes we use a separate index and we define as the first leaf (associated to $\ell_1$) the one on the left.
The \emph{depth} of a tree is the maximum number of splitting nodes visited before reaching a leaf. In general, decision trees need not be binary or complete. However, all decision trees can be transformed into a complete binary decision tree by increasing the depth of the tree and introducing “dummy” splitting nodes. 
Without loss of generality, here we only consider binary decision trees.
A complete binary tree of depth $d$ has $2^d$ leaves and $2^d-1$ splitting nodes. 
\emph{Random forests}, proposed by Leo Breiman~\cite{Breiman2001}, are an ensemble learning algorithm that are based on decision trees.
An ensemble learner incorporates the predictions of multiple models to yield a final consensus prediction. More precisely, a random forest $RF$ consists of $m$ trees, $T_1,\ldots,T_m$, and scoring $RF$ on input $\vec{x}$ means computing $y=\frac{1}{m}\sum_{i=1}^m T_i(\vec{x})$. Let $d$ be the maximum of the depths of the trees in $RF$, we refer to $d$ and $m$ as the \emph{hyperparameters} of the forest.

\paragraph{Polynomial Representation.}
We can represent a tree using polynomials.
Let $T$ be a complete binary tree of depth $d$, then we associate each leaf  with the product of $d$ binomials of the form $(x_i-1)$ or $(x_i+1)$ using the following rule: in the root-leaf path, if at the node $N_i$ we take a left turn we choose $(x_i-1)$, otherwise we choose $(x_i+1)$. We indicate with $P_{d,i}$ the  polynomial of degree $d$ corresponding to the $i$-th leaf. Notice that $P_{d,i}$ contains  only $d$  variables, out of the $2^d-1$ total possible variables (one for each splitting node). We call $\I_{d,i}$ the set of indices of the variables that appears in $P_{d,i}$  and we write $P_{d,i}((x_j)_{j\in\I_i})$ to indicate this; in Fig.~\ref{ex1},  $\I_{2,1}=\I_{2,2}=\{1,2\}$ and $\I_{2,3}=\I_{2,4}=\{1,3\}$. 
Now $T(\vec{x})$ can be computed by evaluating the polynomials $\{P_{d,i}((x_j)_{j\in\I_i})\}_{i=1,\dots,2^d}$ on the values $\{n_j\}_{j=1,\dots,2^d-1}$.
Indeed,  if $i^*$ is the unique value for the index $i$ for which $P_{d,i} ((n_j)_{j\in\I_i}))\neq0$, then $T(\vec{x})=\ell_{i^*}$.

\subsection{Cryptographic Tools}
\label{sec:crypto}
Let $(\M,+)$ be a finite group. A \emph{linearly-homomorphic encryption}  (LHE) scheme for messages in $\M$ is defined by three algorithms: The key-generation algorithm $\gen$ takes as input the security parameter $\kappa$ and outputs the pair of secret and public key, $(\sk, \pk)$. The encryption algorithm $\enc$ is a randomized algorithm that takes in input $\pk$ and $m$ from $\M$, and outputs a  ciphertext, $c \gets \enc_\pk(m)$. The decryption algorithm $\dec$ is a deterministic function that takes as input $\sk$ and $c$, and recovers the original plaintext $m$ with probability $1$ over $\enc$'s random choice.
The standard security property (semantic security) states that it is
infeasible for any computationally bounded algorithm to gain extra
information about a plaintext when given only its ciphertext $c$ and the
public key.  Moreover, we have the homomorphic property: Let
$\C$ be the set of all possible ciphertexts, then there exists an
operation $\odot$ on $\C$ such that for any $a$-tuple of ciphertexts
$c_1 \leftarrow\enc_\pk(m_1), \dotsc, c_a \leftarrow\enc_\pk(m_a)$, it holds that
$\dec_\sk(c_1\odot \cdots \odot c_a) = m_1 + \cdots + m_a$ (with probability $1$).
This implies that, if $c = \enc_\pk(m)$ and $a$ is a positive integer,
$\dec_\sk(\mult(a,c))=am$, where $\mult(a,c)=c\odot\cdots\odot c$ ($a$ times).
Known instantiations of this primitive include Paillier's scheme~\cite{Pal99} and the Joye-Libert scheme~\cite{JL13}.

In the design of the privacy-preserving system presented later on in this work, we will  deploy the LHE-based \emph{secure comparison protocol} described in Fig.~\ref{fig:psc} (protocol $\Psc$).
This is a modification of the protocol presented in \cite{KT06}. Party 1 has the encryption of an integer $a$, while party 2 knows the corresponding secret key. Using the homomorphic property of the encryption scheme, party 1 computes $c'=\enc_{\pk}(\alpha(ra+s))$, where $\alpha$ is sampled uniformly at random from $\{-1,+1\}$ and $r,s$ are integers sampled uniformly at random from $[1,R]$.
Now party 2 receives $c'$ and gets $c=\alpha(ra+s)$. 
It's easy to check that if $0 \leq s < r$, then $\sign(c)=\alpha\,\sign(a)$ (\ie, at the end of the protocol the two parties have multiplicative shares of $\sign(a)$). 
Moreover, if $a$ is an $\ell$-bit integer, then $c$ efficiently hides it when $R=2^{2\ell}$; assuming that $\Z_{2^k}$ is the message space (\eg Joye-Libert scheme) and representing negative integer using the upper half, we can avoid overflow choosing  $k\geq 3\ell$ (see \cite{KT06} for more details). 

\begin{figure}
	\begin{mdframed}
		\begin{itemize}[leftmargin=*,label=--]
			\item \emph{Parties}: Party 1 with input $a'=\enc_\pk(a)$, party 2 with input $\sk$.
			\item \emph{Output}: $\alpha$ for party 1 and $\beta$ for party 2 such that $\alpha\beta=\sign(a)$.
		\end{itemize}
		\begin{enumerate}[leftmargin=*]
			\item Party 1 samples $r\leftarrow[1,R]$, $s\leftarrow[0,r-1]$ and $\alpha\leftarrow\{-1,+1\}$; then it computes $c'=\mult(\alpha r,a')\odot\enc_\pk(\alpha s)$ and send it to party 2. 			
			\item Party 2 computes $c=\dec_\sk(c')$ and $\beta=\sign(c)$.				
		\end{enumerate}	
	\end{mdframed}
\caption{The secure comparison protocol  $\Psc$ that computes the multiplicative sharing of the sign of an encrypted integer.}
\label{fig:psc}	
\end{figure}	

%
	
Finally, a \emph{1-out-of-$n$ Oblivious Transfer} for $c$-bit strings, $\binom{n}{1}$-OT, is a basic two-party cryptographic primitive where party 1 (the sender) has $n$ bit strings, $\{x_1,\dots,x_n\}\subseteq\bit^c$, and party 2 (the receiver) has an index $i\in\{1,\dots,n\}$. At the end of the OT protocol, the receiver has $x_i$ and nothing else about the rest of the sender's messages; the sender learns nothing. For efficient instantiations of this primitive see for example \cite{NP01,ALSZ13,CO15}.

\section{Proposed System}
\label{sec:system}
In this section we describe our system, where the prediction for a new instance  is computed using the random forests trained by different and  mutually distrustful parties on their local data silos. We start by describing the role of the parties involved and  the security model.

\begin{itemize}
	\item \emph{Providers}: There are $t$ providers, the $k$-th one, $\MP_k$, has a random forest $RF_k=\{T^k_1,\dots,T^k_{m_k}\}$ with $m_k$ decision trees; we assume that the forest hyperparameters ($m_k$ and maximum tree depth $d_k$) are public values, while  the description of the trees is the secret input of $\MP_k$ to the system. The providers have no output.
	\item \emph{Server}: The server has no input and no output; it is not trusted to handle private data neither proprietary models\footnote{
		More in general, the proprietary models and the sensitive inputs can not be seen in the clear by the server.}.
	Its function is providing reliable software and hardware to  store encrypted version of the models $RF_k$  and handling prediction request from the user in real-time. 
	\item \emph{User}: Its secret input is an instance $\vec{x}\in\R^n$ ($n$ is public) and the output is the prediction for $\vec{x}$ according to all the trees $T^k_j$; more precisely, the user's output from the system is 
	$$y=\frac{1}{m}\sum_{k=1}^t\sum_{j=1}^{m_k}T^k_j(\vec{x}) \quad \text{ where }  m=m_1+\cdots+m_k.$$

\end{itemize}

We assume that all the parties involved are honest-but-curious (\ie, they always follow the specifications of the protocol but try to learn extra information about other parties secret input from the messages received during the execution of the protocol) and non-colluding (\eg, in real world applications, physical restrictions or economic incentives can be used to assure that the server  has no interest in colluding with another party).
Using the cryptographic tools described in Sect.~\ref{sec:crypto},  we design a system where only the user gets to know $y$ and it gets no other  information about the private models held by the providers. Moreover, the providers and the server gain no information about the input $\vec{x}$. 
We assume that a LHE encryption scheme $(\gen,\enc,\dec)$ is fixed and that there is a party (different form the server) that runs $\gen(\kappa)$, makes $\pk$ public and safely stores $\sk$.  The user needs to authenticate itself with this  party in order to get the secret key $\sk$. Notice that the role of this party can be assumed by the user itself or by one or more of the providers\footnote{
	\eg,  the providers  can use a standard MPC protocol to generate a valid pair $(\pk,\sk)$ in such a way that $\pk$ is public and $\sk$ is secret-shared among all of them. The user  has to “require access” to the system asking to each provider the key-share.}. 
Moreover, we assume that $F:\{0,1\}^\gamma\times\Z^2\rightarrow\M^n$ is a public  pseudorandom function (PRF). 

The system  we present has two phases: an off-line phase, during which each provider  uploads an encrypted form of its forest to the server, and an on-line phase, during which the prediction for a specific input $\vec{x}$ is computed by the server and the user. 
Notice that the off-line phase  is independent of the actual input of the user and  needs to be executed only once (\ie, when the providers join the system). After that, the providers can leave the system and the server will manage each prediction request.  In particular, for each request, a new on-line phase is executed by the server together with the user making the request.  Each phase is described below:

\paragraph{Off-line Phase.} The goal of this phase, which is independent of the actual input of the client,  is to transmit all the trees to the server, but in encrypted form. That is, the server will know the public hyperparameters of each locally learned forest but have no knowledge about the structure of the trees in the forests (\ie, it does not know the indices $i_j$, the thresholds $t_i$, or the leaf values $\ell_i$). This is achieved by having each provider execute the \emph{model-encryption procedure} described in Fig.~\ref{fig:model_enc}. Using this, $\MP_k$ encrypts the thresholds  and the leaf values; moreover it hides the vectors $\vec{e}_{j_i}$ by adding the output of the PRF. At the end,  $\MP_k$ has the encrypted forest ($\Tenc{T^k_j}$ for $j=1\dots,m_k$) and the seed $\vec{s}_k$ used for the PRF. Now, $\MP_k$ sends to the server the encrypted model and makes public an encryption of the seed, $\vec{s}'_k=\enc_\sk(\vec{s}_k)$. 

\paragraph{On-line Phase.} For each prediction request, this phase is executed. A user with input $\vec{x}$ joins the system and sends its  encrypted input to the server. Recall that the latter knows all encrypted forests and the user knows  $\sk$ (the providers are off-line). Now,  the user and the server use this information to run the \emph{tree evaluation protocol $\Pte$} on each encrypted tree $\Tenc{T^k_j}$  (details in Section~\ref{sec:te}).  Protocol $\Pte$ returns an additive sharing of $T^k_j(\vec{x})$, that is  the server and the user get $r^k_j$ and $s^k_j$, respectively and such that $T^k_j(\vec{x})=s^k_j+r^k_j$. In the last step of the on-line phase, the server sends the sum $r$  its shares (one for each tree) and the user computes $y$ as $(s + r)/m$, where $s$ is the sum of the user's shares. See Fig.~\ref{fig:online}

\begin{figure}
\begin{mdframed}
Let $(\gen, \enc, \dec)$ be an LHE scheme with message space $\M$ and $F:\{0,1\}^\gamma\times\Z^2\rightarrow\M^n$ be a PRF.
We assume that $\MP_k$ knows $\pk$, valid public key for the scheme; moreover, $\MP_k$, has a random forest $RF_k=\{T^k_1,\dots,T^k_{m_k}\}$.

\begin{enumerate}[leftmargin=*]
	\item $\MP_k$ samples a seed $\vec{s}_k\leftarrow\{0,1\}^\gamma$  and  completes all the trees in the forest.
	\item Then, for each completed binary tree $T^k_j$, $\MP_k$ does the following:

   \begin{itemize}[leftmargin=*,label=\tiny$\bullet$]
	    \item Assume that the tree has depth $d$ and is represented as described in Section \ref{sec:dt}. That is, each splitting node $N_i$ is identified by the pair $(j_i,t_i)$ and the $i$-th leaf is associated with the value $\ell_i$;
		\item For $i=1,\dots, 2^d-1$, compute  $\vec{e}'_{i}=\vec{e}_{j_i}+ F(\vec{s}_k,(j,i))$ and $t'_i=\enc_\pk(-t_i)$; for $i=1,\dots, 2^d$, compute $\ell'_i=\enc_\pk(\ell_i)$. 
		\item Define $\Tenc{T^k_j}$ (to which we refer as encrypted tree)  as the list  of all values $\big(\vec{e}'_{i},t'_i, \ell'_i,\ell'_{2^d}\big)_{i=1,\dots,2^{d}-1}$.
	\end{itemize}
	\item The  output of the procedure is $\Tenc{T^k_1},\dots,\Tenc{T^k_{m_k}}$ and the  seed $\vec{s}_k$.
\end{enumerate}
\end{mdframed}
\caption{The model-encryption procedure used in the off-line phase of our system.}
\label{fig:model_enc}
\end{figure}

\begin{figure}
\begin{mdframed}
	\begin{itemize}[leftmargin=*,label=--]
		\item\emph{Parties}: The user  with input $\vec{x}\in\R^n$ and the server with  encrypted models $\{\Tenc{T^k_j}_\pk\}_{j=1,\dots,m_k}$, $k=1,\dots,t$.
		\item\emph{Output}: $y$ for the user.\\
	\end{itemize}
	
	We assume that the user has authenticated itself to get $\sk$ and has computed $\vec{s}_k=\dec_\sk(\vec{s}'_k)$ for all $k=1,\dots,t$.
	\begin{enumerate}[leftmargin=*]			
		\item \emph{(Encrypted input submission)} The user computes and sends to the server the values $\vec{x}'[i]=\enc_\pk(\vec{x}[i])$, with $i=1,\dots,n$.
			
		\item \emph{(Tree by tree evaluation)} For $k=1,\dots,t$ and $j=1,\dots,m_k$, the server and the user run the protocol $\Pte$ for the encrypted tree $\Tenc{T^k_j}$. Let $r^k_j$ and $s^k_j$ be the output for the server and the user, respectively.
			
		\item \emph{(Prediction)} The server computes $r=\sum_{k=1}^{t} \sum_{j=1}^{m_k} r^k_j$ and sends it to the user. The latter computes $y=\frac{1}{m}(s+r)$, with $s=\sum_{k=1}^{t} \sum_{j=1}^{m_k} s^k_j$.

		\end{enumerate}
	\end{mdframed}
	\caption{The on-line phase of our system.}
	\label{fig:online}
\end{figure}

\subsection{Tree Evaluation}
\label{sec:te}

\begin{wrapfigure}{R}{0.36\textwidth} 
	\centering
	\begin{tikzpicture}[level distance=1cm,
	level 1/.style={sibling distance=2cm},
	level 2/.style={sibling distance=1cm}, scale=0.8]
	\tikzstyle{every node}=[circle,draw, scale=0.7]
	\node (Root)  {$N_1$}
	child{
		node[fill=green!20] {$N_2$}  
		child { node[fill=green!60] {$\ell_1$} edge from parent node[left,draw=none] {$-1$}}
		child { node[fill=green!80!blue] {$\ell_2$} edge from parent node[right,draw=none] {$+1$}}
		edge from parent node[left,draw=none] {$+1$}	
	}
	child{
		node[fill=orange!40] {$N_3$}
		child { node[fill=orange!80!red] {$\ell_3$} edge from parent node[left,draw=none] {$+1$}}
		child { node[fill=orange!70] {$\ell_4$} edge from parent node[right,draw=none] {$-1$}}
		edge from parent node[right,draw=none] {$-1$}
	};	
	\end{tikzpicture}\\[.2cm]
	$\begin{aligned}
	P^\gamma_{2,1}(x_1,x_2)=(x_1+1)(x_2-1)\\
	P^\gamma_{2,2}(x_1,x_2)=(x_1+1)(x_2+1)\\
	P^\gamma_{2,3}(x_1,x_3)=(x_1-1)(x_3+1)\\
	P^\gamma_{2,4}(x_1,x_3)=(x_1-1)(x_3-1)
	\end{aligned}$	
	\label{ex:rand}
	\caption{Randomized representation  (with $\gamma_1=-1$, $\gamma_2=+1$, $\gamma_3=-1$) for the tree in Fig.~\ref{ex1}. 
		The light-green node, which is the 3rd splitting node in the original tree, becomes the 2nd splitting node.}	
\end{wrapfigure}

In protocol $\Pte$ (Fig.~\ref{fig:pte}) the server and the user compute an additive sharing of $T^k_j(\vec{x})$ from $\Tenc{T^k_j}$ and $\vec{x}'=\enc_\pk(\vec{x})$.
Recall from Sect.~\ref{sec:dt} that, given a tree $T$ and an input $\vec{x}$, finding the index $i^*$  such that the polynomial $P_{d,i^*}$ evaluates $0$ on the values $\{n_j\}_j$ is equivalent to compute $T(\vec{x})$ (\ie, $T(\vec{x})=\ell_{i^*}$).
Therefore, finding $i^*$  is sufficient in order to then compute an additive sharing of $T(\vec{x})$. In the privacy-preserving scenario, the main challenges in doing this are: 1) First of all, notice that neither the server or the user can see $i^*$ in the clear, indeed knowing  the index of the reached leaf can leak information about the inputs and the tree structure (when more than a request is made). We solve this using a simple \emph{tree randomization} trick that  hides $i^*$ for the user 
(\ie, the user gets to know $i^*$ for an tree $T'$ equivalent to $T$ but with nodes randomly permuted by the server) 
and an OT-channel that hides $i^*$ for the server (\ie, once that the user gets $i^*$ for $T'$, the oblivious transfer protocols  allows it to receive $\ell_{i^*}$  without revealing $i^*$ to the server). 2) Then observe that neither the server or the user can see the  $\{n_j\}_j$ in the clear, indeed  also these values can leak information about $\vec{x}$ or $T$. To solve this we use the homomorphic property of the underlying LHE to let the server compute encryption of each $N_j(\vec{x})$, the secure comparison protocol $\Psc$ to share $n_j=\sign(N_j(\vec{x}))$ between the server and the user, and finally an algebraic property of the polynomials $\{P_{d,i}\}_i$ to let the user compute $i^*$ from its local shares of $\{n_j\}_j$.

More in details, in step 1 for each splitting node, the server computes $N_i'$, an encryption of $N_i(\vec{x})$, using the homomorphic properties of the underlying scheme and the formula $N_i(\vec{x})={\vec{e}'}_i^\tr\cdot\vec{x}-t_i-b_i$. The value $b_i$ is computed by the user using the knowledge of the seed $\vec{s}_k$.
In step 2 (tree randomization), the server computes a randomized  representation in order to hide from the user the path corresponding to traversing the tree $T^k_j$ on input  $\vec{x}$. 
Recall the standard polynomial representation described in Section~\ref{sec:dt}, such representation can be randomized in the following way: For each splitting node $N_i$, sample $\gamma_i\in\{-1,+1\}$; if $\gamma_i=-1$ swap the left and the right subtrees of $N_i$ and proceed to re-index the splitting nodes and the leaves according to the standard indexing system described in Sect.~\ref{sec:dt}. 
Let  $\gamma=(\gamma_1,\dots,\gamma_{2^d-1})\in\{-1,+1\}^{2^d-1}$, 
now the root-leaf path for the $i$-th leaf is represented by the polynomial
$P_{d,i}^\gamma$  obtained by taking $P_{d,i}$ and changing the $j$-th factor  from $x_{i_j} - 1$ (or $x_{i_j} + 1$) in $x_{i_j} -\gamma_{i_j}$ (or $x_{i_j} +\gamma_{i_j}$), see Fig.~\ref{ex:rand}.
Notably, it holds that $T^k_j(\vec{x})=\ell_{i^*}$ where $i^*$ is the unique value for the index $i$ such that $P^\gamma_{d,i} ((n_j)_{j\in\I_i}))\neq0$.
In step 3,  the two parties use the secure comparison protocol $\Psc$ of Sect.~\ref{sec:crypto} in order to compute multiplicative shares of $n_i$ from $N_i'$.  In particular, the server and the  know $\alpha_i$ and $\beta_i$, respectively and such that $\alpha_i\beta_i=n_i$.
Now, notice the following: let $\tilde{\gamma}=(\gamma_1\alpha_1,\dots,\gamma_{2^d-1}\alpha_{2^d-1})$, then for each $i=1,\dots, 2^d$ we have
$$
P_{d,i}^{\tilde{\gamma}}((\beta_j)_{j\in\I_i})\prod_{j\in\I_i}\alpha_j=P_{d,i}^\gamma((n_j)_{j\in\I_i}).
$$
For example, $(\beta_1-\gamma_1\alpha_1)(\beta_2+\gamma_2\alpha_2)\alpha_1\alpha_2=(\alpha_1\beta_1-\gamma_1\alpha_1^2)(\alpha_2\beta_2+\gamma_2\alpha_2^2)=(n_1-\gamma_1)(n_2+\gamma_2)$.
Moreover, $P_{d,i}^\gamma((n_j)_{j\in\I_i})=0$ if and only if $P_{d,i}^{\tilde{\gamma}}((\beta_j)_{j\in\I_i})=0 $ and the latter values can be computed locally by the user (after having received $\tilde{\gamma}$ from  the server). Using this, the user computes $i^*$ such that $T^k_j(\vec{x})=\ell_{i^*}$ (step 4).
Finally, in step 5 the servers run a $\binom{2^d}{1}$-OT protocol that allows the user to receive the additive share of $\ell_{i^*}$ without revealing any extra info.

\subsection{Outline of Security Analysis}
\label{sec:security}
Assuming that the OT and the comparison protocol are secure, the security of our system  against a honest-but-curious (non-colluding) server follows from the security of the encryption scheme and PRF used (\ie, we can use the composition theorem~\cite{Canetti00},\cite[Section 7.3.1]{Goldreich2004}). Indeed all the messages received by the server are ciphertexts or values that has been additively masked using the PRF.
The security against a honest-but-curious (non-colluding) user follows from the tree randomization trick. Since a  randomized representation is used for the tree, even if the user gets to know the index $i^*$ of the leaf corresponding to the prediction, such information does not reveal the root-leaf path in the original tree (\ie, the bit-string $\tilde{\gamma}$ received in Step 4 is random). Moreover, the user does not learn any extra information about the local models since it does not see the individual predictions (\ie, the user only see one share $s^k_j$ for each prediction and the distribution of $s^k_j$  is independent of the prediction $T^k_j(\vec{x})$).

\subsection{Complexity}
\label{sec:complexity}
We detail the complexity of our system in terms of cryptographic operations. During the off-line phase, the providers run the model-encryption procedure to encrypt their models $RF_1,\dotsm, RF_t$.
Assume that $RF_k$ has hyperparameters $d_k$ and $m_k$, then for $\MP_k$ the model-encryption procedure costs $\Theta(m_k\,2^{d_k})$ operations (\ie, encryptions and calls to a PRF).
Moreover, $\MP_k$ sends to the server $m_k \,2^{d_k+1}$ ciphertexts and $n\,m_k \,2^{d_k}$ messages from $\M$. 
The complexity of the on-line phase  is dominated by  step 2, where the protocol $\Pte$ is executed $m$ times, one for each tree ($m=m_1+\dots+m_t$). 
Protocol $\Pte$ for a tree of depth $d$ and an instance with $n$ features requires $\Theta(n\,2^d)$ operations ($\Theta(2^d)$ for the user and $\Theta(n\,2^d)$ for the server) and generates $\Theta(2^d)$ ciphertexts exchanged among the server and the user (see Table~\ref{tab:complexity} in the appendix for more details).
Therefore, the on-line phase has complexity proportional to $n\,m\,2^d$, where $d=\max_k\,d_k$.
Finally, notice that many steps of our system can be easily \emph{run in parallel}. For example, the $m$ needed instances of protocol $\Pte$ can be executed concurrently.  


\begin{figure}
	\begin{mdframed}
		\begin{itemize}[leftmargin=*,label=--]
			\item \emph{Parties}: The server, with encrypted tree $\Tenc{T^k_j}$ and encrypted input $\vec{x}'$, and the user with $\sk$. The depth of $T^k_j$ is $d$.
			\item \emph{Output}: $r^k_{j}$ for the server, $s^k_j=T^k_j(\vec{x})-r^k_{j}$ for the user.
		\end{itemize}
		\begin{enumerate}
			\item \emph{($N_i(\vec{x})$ computation)} For $i=1,\dots,2^d-1$, the user computes $b_i=F(\vec{s}_k,(j,i))^T\cdot \vec{x}$ and $b'_i=\enc_\pk(-b_i)$. The values $b'_i$'s are sent to the server, who computes
			$$N'_i=\bigodot_{s=1}^{n}\mult\big(\vec{e}'_{i}[s],\vec{x}'[s]\big)\odot  t'_{i}\odot  b'_{i}.$$

			\item \emph{(Tree randomization)} The server samples $\gamma_i\leftarrow\{-1,+1\}$ and proceeds to the tree randomization. From now on, the randomized tree is used.
			
			\item \emph{(Node evaluation)} At this point $N_i'=\enc_\pk(N_{i}(\vec{x}))$), the server and the user run protocol $\Psc$ where the server is party 1 with input $N'_i$ and the user is party 2. Server gets $\alpha_i$ and user gets $\beta_i$ ($i=1,\dots,2^d-1$).
			
			\item \emph{(Path computation)} Define $\tilde{\gamma}_i=\gamma_i\alpha_i$, the server sends $\tilde{\gamma}=(\tilde{\gamma}_1,\dots,\tilde{\gamma}_{2^d-1})$ to the user, who computes 
			$z_i=P_{d,i}^{\tilde{\gamma}}((\beta_j)_{j\in\I_i})$ for all $i=1,\dots,2^d$. Let $i^*$ be the index such that $z_{i^*}\neq0$.
			
			\item \emph{(Label computation)}  The server samples  $r^k_{j}\leftarrow\M$ and computes $\tilde{\ell}_i=\ell'_i\odot\enc_\pk(-r^k_{j})$ for all $i=1,\dots,2^d$. 
			Now server and user  run a $\binom{2^d}{1}$-OT where the server  is the sender with $\tilde{\ell}_1,\dots,\tilde{\ell}_{2^d}$ as input and the user is the receiver with input $i^*$.  Let $c^k_j$ be the ciphertext received by  the user, then the latter computes $s^k_j=\dec_\sk(c^k_j)$.
		\end{enumerate}
	\end{mdframed}
	\caption{The tree-evaluation procedure (protocol $\Pte$) used in the on-line phase.}
	\label{fig:pte}
\end{figure}

\section{Random Forest Hyperparameters}
\label{sec:hyper}

Since the  depth and number of trees (\ie model hyperparameters) affect the efficiency of our system (see Sect.~\ref{sec:complexity}), we provide here an empirical demonstration that bounding them can be done without adversely  the prediction efficacy.

\begin{wrapfigure}{R}{0.65\textwidth} 
	\centering
	\includegraphics[width=.6\textwidth]{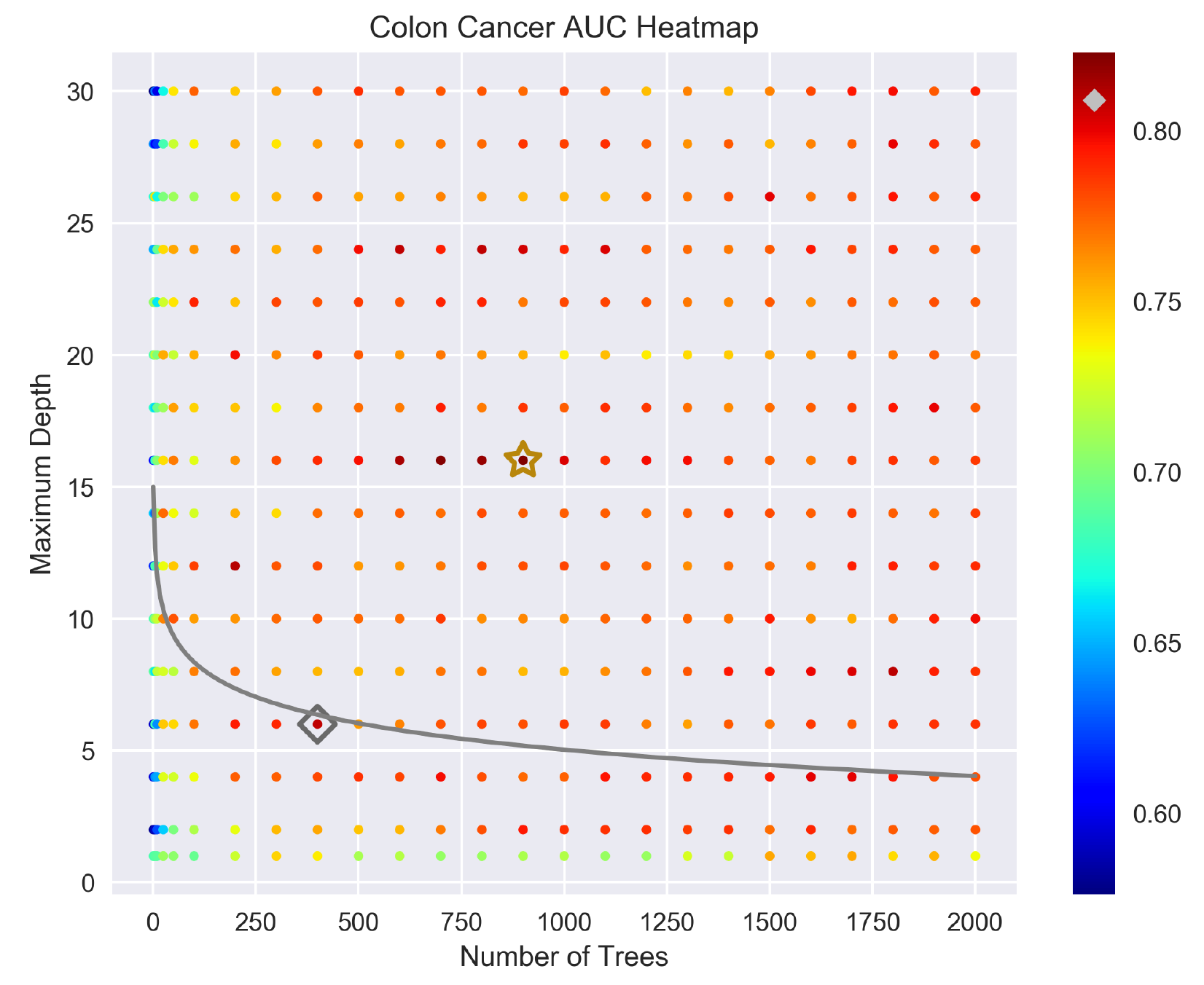}
	\caption{
		Heatmap of mean AUC values for various combinations of $d$ and $m$. The  gray line indicates $m\,2^d=2^{15}$. The gold star indicates the best overall combination of $d$ and $m$ (AUC=0.823), the silver diamond indicates the best overall combination  constrained by $m\,2^d\leq2^{15}$ (AUC=0.809). The silver diamond is also on the colorbar indicating the corresponding AUC.}
	\label{fig:heatmap}
\end{wrapfigure}

\paragraph{Bounded Depth.}
Typically, during the training phase a random forest is grown such that each tree may split in a greedy fashion without concern for the depth of the trees.
We provide here an empirical inspection of the effect of bounding the depth to a maximum value $d$ on the efficacy (AUC value\footnote{With labels $\{+,-\}$, the AUC (area under the curve) is the probability that a model classifies a randomly chosen $+$ instance higher than a randomly chosen $-$ one.}) of the learned forest. We utilize the public Kent Ridge Colon-cancer dataset from the UCI repository (reference in Table~\ref{tab:UCI}) and we looked at various combinations of $d$ and the number of trees in the forest, $m$.
Specifically, we consider values of $d$ in $\{1, 2, \ldots, 28, 30\}$ and 25 different choices of $m$ in $\{1, 5, 10, 25, 50, 100, 200, 300, \ldots, 1900, 2000\}$.
For each  pair of values, we performed 30 replicates of a random forest model construction and evaluation process.
For each model, the construction began with choosing a random 70\% of the data to serve as training data and the remaining 30\% as testing data.
A model was then built with the specified hyperparamters and AUC was measured on the testing data.
In Fig.~\ref{fig:heatmap} we present the results of this investigation as a heatmap. For this task even a maximum depth of 6 was competitive with larger depth choices if 300 trees are considered.
This suggests that while the standard learning algorithm may greedily grow trees very deeply, the overall performance is not substantially impacted by bounding the maximum depth.

\paragraph{Tuning Methodology.} 
Common practice for training ML algorithms involves some selection method for determining a choice for the hyperparameters.
One standard selection method is a grid-based search wherein a researcher will predefine some set of choices for each hyperparameter  and then the cross product of these sets and choose the combination that maximized the AUC of the model. For example, for random forest, we pick the hyperparameters as $d^*, m^* = \argmax_{(d, m) \in D \times M} AUC(RF(d, m))$, where $RF(d, m)$ is a random forest trained with hyperparameters $d$ and $m$, $AUC(\cdot)$ is the AUC of a given random forest on some held aside validation data and $D,M$ are fixed sets. However, this procedure searches all combinations of $d$ and $m$, whereas we are interested in controlling the value $m\,2^d$ because the overhead of our system its directly proportional to it in Sect.\ref{sec:complexity}. Therefore, between two hyperparameters choices giving the same efficacy, we are interested to choose the one that produces smaller overhead. 
In other words, our approach for tuning is the following: we fix a  value $s$ and then we maximize the model efficacy constrained to choosing the hyperparameters $d$ and $m$  in the set  $Q_s = \{(m, d) \in \mathbb{Z}^+ \times \mathbb{Z}^+ \mid m\, 2^d \leq s\}$. 
The gray line in Fig.~\ref{fig:heatmap} depicts the boundary of $Q_s$ when $s=2^{15}$ and dictates that choices above it are too large, and choices below are of acceptable overhead.
Even if the number of acceptable choices  is relatively small compared to the total  number of combinations, it is worth noting that we saw competitive performance as both depth and number of trees exceeded some minimum choices. This suggests that we may be able to achieve both good performance and small  overhead.

\section{Performance}
\label{sec:performance}

\begin{wrapfigure}[31]{R}{0.65\textwidth}
	\centering
	\includegraphics[width=.6\textwidth]{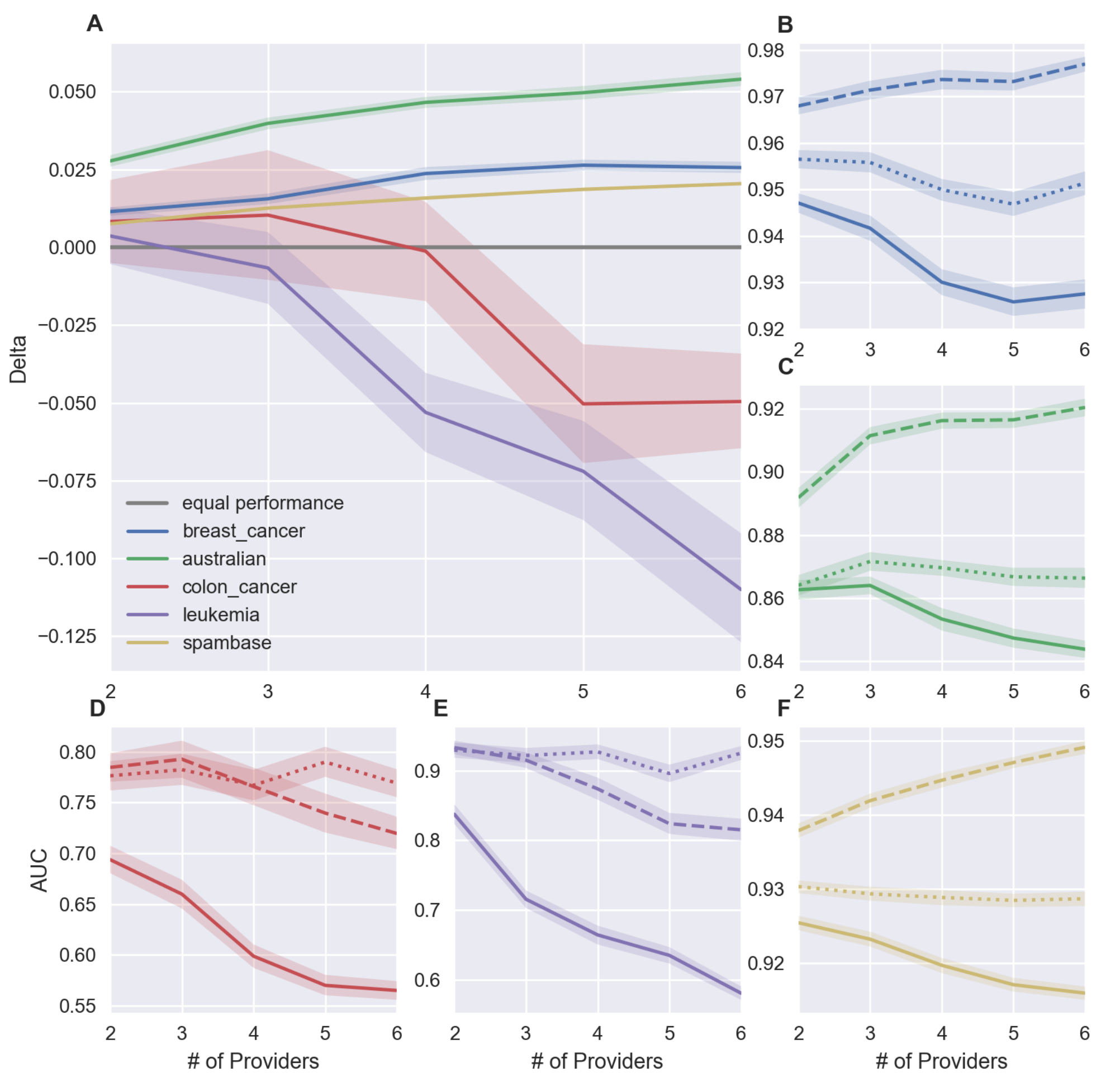}
	\caption{
		Effect of locally learning then merging compared to learning from a merged dataset.
		Subfigures \textbf{B}-\textbf{F} shows on the datasets of Table~\ref{tab:UCI} how AUC is impacted by the number of providers. The dashed, solid and dotted lines shows AUC values for the locally learn then merge, the zero-sharing and the merge then learn approach, respectively. Subfigure \textbf{A} shows the AUC difference between the locally learn then merge and merge then learn (positive values indicate an improvement using our approach).
	} \label{fig:merge}
\end{wrapfigure}

To conclude our work, we want to experimentally validate our system.

\subsection{Efficacy} 
First,  we study the effect of the \qquote{locally learn then merge} approach on the prediction accuracy.  In particular, we want to compare the accuracy of the proposed method with the one of the standard \qquote{merge then learn} approach\footnote{If there are no privacy concerns, parties can simply share their data with one another and learn a single model. 
Otherwise a privacy-preserving training algorithm can be used to achieve the same result.}.
We provide an empirical investigation of this in two fashions:  across three disease prediction tasks using EHR data from the Marshfield Clinic in Marshfield, WI, and across five predictions tasks from the UCI database.

\paragraph{Real EHR Data.}
We consider the tasks of predicting three diseases 1 month in advance: Influenza (ICD-9 487.1), Acute Myocardial Infarction (ICD-9 410.4), and Lung Cancer (ICD-9 162.9).
Each dataset was comprised of up to 10,000 randomly selected case-control matched patients on age and date of birth (within 30 days), with cases having 2 or more positive entries of the target diagnosis on their record and the control having no entries (rule of 2).
Data for each case-control pair were truncated following 30-days prior to the case patient's first diagnosis entry to control for class-label leakage.
Features were comprised of patient demographics, diagnoses, procedures, laboratory values and vitals.
Unsupervised feature selection was performed on a per-disease basis first with a 1\% frequency-based filter to remove very uncommon features and then followed up with principal component analysis to reduce the overall dimension of the data to 1,000 (this was done to improve the performance speed of our algorithm).
For each of the 3 diseases, we constructed, as a performance baseline, a random forest model with 500 trees, a maximum depth of 8, and 10\% of features considered at each split in the tree.
Models were trained on 90\% of the data and tested on a held aside 10\%.
We compared these baseline models (\ie, \qquote{merge then learn} approach) with our \qquote{locally learn then merge} approach by again constructing a forest with the same hyperparameters except the training data were partitioned between two simulated providers each with 45\% of the original data that were used to train two smaller forests of 250 trees each and then merged together.
Model performance was measured using the area under the receiver operating characteristic curve (AUC), a common ML accuracy metric.
We present in Table \ref{tab:eff_res} both the dataset information and results of our experimentation.
We find that the performance of our \qquote{locally learn then merge} AUC values are comparable across all three prediction tasks.

\begin{table}
	\centering
	\renewcommand{\arraystretch}{1.1}	
	\begin{tabular}{|l|l|l|l|l|l||l|}
		\hline
		ICD-9 & Disease     & Samples & Features & Base AUC & LLM AUC & Prediction Time (s)\\
		\hline
		487.1 & Influenza   & 10,000  & 8,211    & 0.8011   & 0.7640  & 105.37$\pm$14.70   \\
		410.4 & Acute MI    & 9,284   & 9,136    & 0.6797   & 0.6658  & 121.75$\pm$9.43   \\
		162.9 & Lung Cancer & 10,000  & 9,021    & 0.6313   & 0.5786  & 125.94$\pm$8.19  \\\hline      
	\end{tabular}
	\caption{Efficacy testing results for 3 EHR datasets.
		Number of features are calculated before applying PCA (post-PCA selected the top 1,000 components).
		Base AUC refers to a forest learned on the whole dataset (\qquote{merge then learn} approach) and LLM AUC refers to a forest learned in our \qquote{locally learn then merge} fashion.
		Prediction Time refers to the mean$\pm$std time required for our system to return a prediction for a single patient query.
	}
	\label{tab:eff_res}
\end{table}

\paragraph{UCI datasets.}
We use five \href{https://archive.ics.uci.edu/ml/datasets.html}{UCI} datasets (references in Table~\ref{tab:UCI}) to investigate the effect of the number of providers sharing data on the performance of a random forest. To simulate a dataset being shared amongst $t$ providers, we randomly split each UCI dataset into $t$ equal sized and unique chunks, $D_1, \dots, D_t$, with each chunk belonging to a single provider. Each chunk was then split into a training (70\% of the data) and testing set (30\% of the data), \ie $D_i = \textup{Train}_i \cup  \textup{Test}_i$.
We then learned models in three different ways.
To simulate the effect of \qquote{zero sharing} (\ie, providers with silo data do  not share data or models), provider $i$ learns a forest on $ \textup{Train}_i$ and tests on $ \textup{Test}_i$ achieving $AUC_i$ with the average silo AUC taken as the mean across all $t$ providers. 
Each forest was learned with 50 trees of maximum depth 8.
To simulate the effect of \qquote{locally learn then merge}, each provider  learns a random forest on their own training data, the forests are merged together, and the AUC is calculated on the merged testing data, $\cup_i  \textup{Test}_i$.
Again, each provider learned 50 trees of maximum depth 8 and the final merged forest being of size $50\,t$ trees.
To simulate the effect of a merged dataset (\qquote{merge the learn}) we learn a single forest with $50\,t$ trees and maximum depth 8 from  $\cup_i  \textup{Train}_i$ and then evaluate the AUC on  $\cup_i  \textup{Test}_i$. 
This process was repeated 50 times to produce confidence intervals and performed for each of the five datasets in Table~\ref{tab:UCI} across five choices of $t \in \{2, 3, 4, 5, 6\}$. 
We present the results of these experiments in Fig.~\ref{fig:merge}. 
We see from it that the effect of locally learning the merging has neither a strictly positive or negative effect on the quality of the model. Indeed,
our results indicate that the effect is dataset dependent. Therefore, we believe that it would be critical for a provider   to investigate how the quality of their predictions are impacted by merging their learned models with another hospital system as compared to using their own data.

\subsection{Efficiency}
\label{ssec:efficency}

\begin{wraptable}{R}{0.55\textwidth} 
	\centering
	\renewcommand{\arraystretch}{1.1}  	
	\begin{tabular}{|l|cc|cc|}
		\hline
		& size & $n$ &  \multicolumn{2}{c|}{Model-encryption}  \\ 
		& & &                                           Time ($s$)           & Size (MB)          \\ 
		\hline 
		Australian & 609 & 14 	& 0.84 & 7.96\\
		Breast cancer & 569 & 30 & 0.90   & 9.6\\
		Spambase & 4601 & 57 & 1.01  & 12.35\\
		Colon cancer  & 62 & 2000 & 10.57 & 210.54  \\
		Leukemia & 72 & 7129 & 41.7 & 733.69 \\
		\hline
	\end{tabular}
	\caption{References for the UCI datasets ($n=$ number of features). The last two columns show the overhead of the off-line phase of our system.}
	\label{tab:UCI}
\end{wraptable}

\paragraph{Implementation details.}
To test efficiency (\ie, bandwidth and running time) we implemented our proposed system  in Python3.5.2.
As underlying LHE we use  Joye-Libert's scheme~\cite{JL13} with $\M=\Z_{2^{64}}$ and  100-bit security. We assume  all inputs are real number with absolute value less or equal to $2\cdot10^3$ and at most 3 digits in the fractional part. To convert them into values in $\M$, we multiply each value by $10^3$. This allows to represent all inputs with $21$-bits values (we represent negative values using the upper half of $\Z_{2^{21}}$) and avoid overflow in the secure comparison protocol.
The $\binom{2^d}{1}$-OT protocol for $20148$-bit strings is  implemented~\cite{NP99} using $d$ calls to a  standard $\binom{2}{1}$-OT protocol (\ie, \href{https://github.com/emp-toolkit}{emp-toolkit}) for $100$-bit strings and $2d$ calls to a PRF (\ie, AES$_{128}$).
We provide an empirical investigation of the efficiency in two fashions: using a commodity machine and using the \href{https://research.cs.wisc.edu/htcondor/index.html}{HTCondor system}.

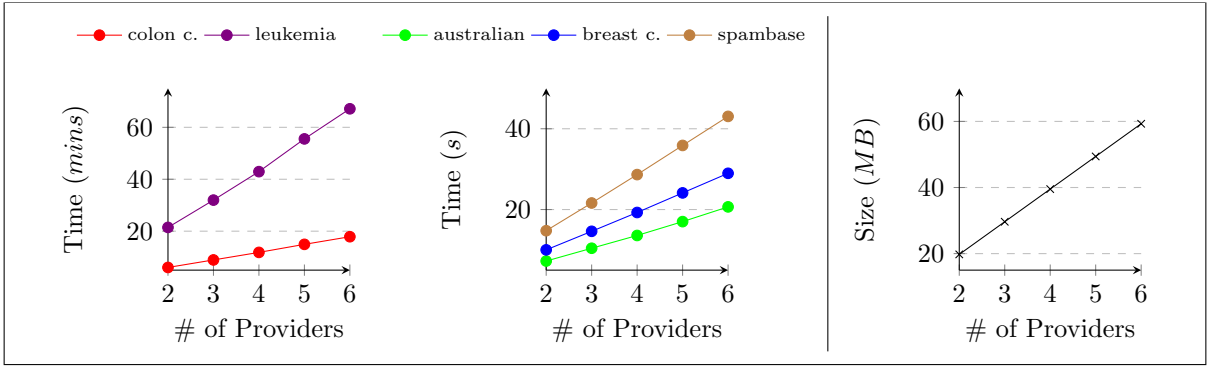
\begin{figure}
\begin{mdframed}
	\centering
		\begin{tikzpicture}
		\begin{axis}
		[
		xmin=2, xmax=6, ymin=5, ymax=75,
		height=4cm, width=4cm,
		ylabel={Time ($mins$)},
		axis lines=left,
		xlabel={\# of Providers},
		legend columns=-1,
		legend style={font=\scriptsize, draw=none, at={(1,1.4)}},
		ymajorgrids=true,
		grid style=dashed
		]
		\addplot[mark=*, red] table[x=t, y=time, col sep=comma] {contents/csv/ccM.csv};
		\addplot[mark=*, violet] table[x=t, y=time, col sep=comma] {contents/csv/leuM.csv};  
		\legend{colon c., leukemia}  
		\end{axis}
		\end{tikzpicture}
		\begin{tikzpicture}
		\begin{axis}
		[
		xmin=2, xmax=6, ymin=5, ymax=50,
		height=4cm, width=4cm,
		ylabel={Time ($s$)},
		axis lines=left,
		xlabel={\# of Providers},
		legend columns=-1,
		legend style={font=\scriptsize, draw=none, at={(1.5,1.4)}},
		ymajorgrids=true,
		grid style=dashed
		]
		\addplot[mark=*, green] table[x=t, y=time, col sep=comma] {contents/csv/australian.csv};
		\addplot[mark=*, blue] table[x=t, y=time, col sep=comma] {contents/csv/bc.csv};
		\addplot[mark=*, brown] table[x=t, y=time, col sep=comma] {contents/csv/spam.csv};
		\addplot[mark=*, red] table[x=t, y=time, col sep=comma] {contents/csv/cc.csv};
		\addplot[mark=*, violet] table[x=t, y=time, col sep=comma] {contents/csv/leu.csv};
		\legend{australian, breast c., spambase}    
		\end{axis}
		\end{tikzpicture}
		\vrule 
		\hspace*{.1cm}
		\begin{tikzpicture}
		\begin{axis}
		[
		xmin=2, xmax=6, ymin=15, ymax=70,
		height=4cm, width=4cm,
		ylabel={Size ($MB$)},
		axis lines=left,
		xlabel={\# of Providers},
		legend cell align={left},
		ymajorgrids=true,
		grid style=dashed
		]
		\addplot[mark=x, black] table[x=t, y=comm, col sep=comma] {contents/csv/ccM.csv};
		\end{axis}
		\end{tikzpicture}
\end{mdframed}
\caption{Performance of protocol $\Pte$ on $(50\times \#$ of Providers) trees of depth $8$ for the datasets of Table~\ref{tab:UCI}.}
\label{fig:servers} 
\end{figure}

\paragraph{Commodity machine.} We report the performance of our system executed on a commodity machine (60GB memory and 48core CPU, Intel Xeon CPU E5-2680 v3) for the UCI datasets of Table~\ref{tab:UCI} in the setting described before (\ie each provider knows a random forest with 50 trees of maximum depth 8).
Several tasks in the implementation were parallelized by multi-threading;
all the timing values are averaged on five repetitions of the same experiment.
Table~\ref{tab:UCI} (last two columns on the right) reports the running time of the model-encryption procedure executed by one provider during the \emph{off-line phase} (Fig.~\ref{fig:model_enc}); it also reports the size of the encrypted model obtained via this procedure. The number $n$ of features influences both results, however even for the high dimensional cases (\ie, thousands of features) the encrypted model has size less than $1$~GB and is produced in less than a minute.
The \emph{on-line phase} of our system consists of three steps (Fig.~\ref{fig:online}): In step 1, the user submits its encrypted input to the server. Clearly, the performance of this step is influenced only by the encryption scheme used and by the dimension of the input (\ie, number of features $n$). In our experiments, even  for the largest value of $n$,  this step takes less than a second (\eg, $0.17$ seconds for $n=7129$). In step 2, the server and the user execute $m$ times the protocol $\Pte$ to evaluate each tree in the merge of all the forests. Fig.~\ref{fig:servers} illustrates the performance of this part (the most expensive one in the on-line phase): The two graphs on the left depict the running time of the protocol $\Pte$ run on $50\, t$ trees as function of the parameter $t$, number of providers; the results are dataset dependent since the server  executes  $\Theta(n\,2^d)$ cryptographic operations. The graph on the right of Fig.~\ref{fig:servers} reports the size of the messages exchanged by the server and the user as function of $t$. This value is not influenced by $n$ (dataset size) and it only increases linearly with the number of trees; in our experiment, even for 300 trees the bandwidth required is always less than $60$~MB. 
In the last step of the on-line phase (step 3), the server and the user sum their shares; the overhead of this step is independent of $n$ and influenced only by the total number of trees (\eg, in our experiment this needs less than $8$~$ms$ for $300$ trees).

\paragraph{HTCondor.}  The experiments for the real EHR data were executed using the HTCondor system, a high-throughput computing architecture that we utilized in a \qquote{master-worker} fashion.
For each forest, one tree was learned as a separate \qquote{job} exploiting the heavy parallelization available to random forests.
Thus, both training and prediction were performed in a high-throughput manner.
We report the running time of the on-line phase in this setting in the last column on the right of Table~\ref{tab:eff_res}.
We find that this parallelized version of our algorithm allows us to provide near real-time interactions as predictions are returned on average within two minutes of providing a query to the system.
We believe that this would be reasonably fast enough to support the workflow of a physician who wishes to query the model for a single patient.

\section{Related Works}
Privacy-preserving training considers the scenario where different parties have their own private datasets and want to train a machine learning model on the merge without each part having to reveal extra information about its local dataset to the others. There is an extensive research that propose  privacy-preserving training systems (see \cite{AggarwalY08a} or \cite{Lindell09} as surveys and \cite{BCMOPS18} for a recent example) and few of them focus on training decision tree. In 2000, Lindell and Pinkas \cite{LP00} presented a system for two parties having each one a private database (both with same feature types) and willing to collaborate in order to run the ID3 learning algorithm on the union of them. Their system uses oblivious polynomial evaluation \cite{NP99} and Yao's protocol \cite{Yao86}. 
Threshold homomorphic encryption is instead used in \cite{XHLS05} and \cite{SM08} in order to consider the case of more than two parties. 
While the former works consider horizontally partitioned data, another line of work assumes data that are vertically partitioned among two or more data holders, see for example \cite{VCKP08}. To the best of our knowledge, de Hoogh et al.~\cite{HSCA14} presented the most efficient protocol for privacy-preserving training of decision trees. Their protocol works for $n\geq 3$ providers with honest majority (\eg, no more than $n/2$ providers collude), is based on Shamir's secret sharing~\cite{Sha79} and is designed for categorical attributes. 
Lastly, in 2014 Vaidya et al.\ \cite{VaidyaSFML14} proposed a method to learn and score \emph{random trees} in a privacy preserving manner.  Like our approach, their approach requires encryption-based collaboration to make predictions.  Unlike our approach, their approach also requires interaction and collaboration at training time.  One party proposes a random tree structure, and all parties must contribute information to the distributions at the leaf nodes.  In our approach, learning is completely independent for each party, and hence training is much faster.  An advantage of Vaidya et al.\ is the ability to also address vertically partitioned data.

The setting where one party, the provider, holds a decision tree and the user holds an instance was first considered in 2007 by Ishai and Paskin \cite{IP07} (privacy-preserving scoring). In the same year, Brickell et al.~\cite{BPSW07}  designed a protocol for evaluating binary branching programs. Two years later, Barni et al.~\cite{BFKLSS09}  improved  the previous result and  extended to linear branching programs.
In 2015, Bost et al.~\cite{BPTG15} used Yao's protocol and levelled homomorphic encryption to design a scoring algorithm for decision trees assuming that the number of nodes is equal to the number of features; in their work they also considered Naive Bayes and hyperplane classification. In 2017, Backes et al.~\cite{BBBEHHL17} improved and extended to random forests the algorithm presented by \cite{BPSW07}. Later on, Wu et al.\ \cite{WFNL16} first and then Tai et al.\ \cite{TMZC17} improved the former result designing more efficient algorithms. In particular, the scoring algorithm of \cite{TMZC17} uses only linearly-homomorphic encryption and is especially efficient for deep but sparse trees.  In 2018, Joye and Salehi \cite{JS18} proposed a protocol that reduces the workload for the provider and server, respect to \cite{WFNL16,TMZC17}. The bandwidth usage for the user is also improved, while for the server the saving depends on the tree sparsity. 
Finally, De Cock et al.\ \cite{DCDHKNPT17} use secret-sharing based protocol to design scoring algorithm for decision trees, SVMs and logistic regression models.

Another line of research focuses on constructing \emph{differentially private decision tree classifiers}, see for example the work of Jagannathan et al.\cite{JPW12} and Rana et al.\cite{RGV15} Our approach is orthogonal to differential privacy since we consider a different threat model.



\section*{Conclusions}
We propose a new approach for computing privacy-preserving collaborative predictions using random forests. Instead of a system composed by a training algorithm, which usually has high overhead in the privacy-preserving setting, followed by a scoring algorithm, we propose a system based on locally learning and then  privacy-preserving merging. To avoid the need for providers to be on-line for each prediction request, we instantiate the new approach in the cloud model. That is, an untrusted server collects the locally trained models in encrypted form and takes care of scoring them on a new private instance held by the user. Our system is secure in the honest-but-curious security model and extending it  to the malicious model, especially for a corrupted server, is an interesting direction for future work.  We evaluate the performance of our system on real-world datasets, the experiments we conducted show that (1) the efficacy of the new approach is dataset dependent;  this opens to future works that aim to characterize this dependency in terms of the dataset parameters and distribution, (2) the efficiency is influenced by the forest hyperparameters, which we showed we can control,  and  by the  number of features $n$, which is given by the specific application; avoiding the dependency on $n$ is another interesting direction that may lead more efficient  implementation of this new approach.

\subsubsection*{Acknowledgments.}
This work was partially supported by the Clinical and Translational Science Award (CTSA) program, through the NIH National Center for Advancing Translational Sciences (NCATS) grant UL1TR002373,  by the NIH BD2K Initiative grant U54 AI117924 and by the NLM training grant 5T15LM007359.

\bibliographystyle{abbrv}
\bibliography{bib}

\appendix
\section{Appendix}
\subsection{Privacy-Preserving Scoring of Random Forests}
\label{app:pps}

In this section we describe a simplified version of the system of  Sect.~\ref{sec:system} that can be used as a standard privacy-preserving scoring algorithm for random forests. In this scenario, we have the following two parties:
The \emph{provider} $\MP$ has a random forest $RF$, and wants to keep it private but it is willing to let clients use it for prediction. The \emph{client} $\cli$ has $\vec{x}$ and wishes to know $RF(\vec{x})$ without revealing $\vec{x}$.
In order to achieve this, we present the protocol $\Ppps$ in Fig.~\ref{fig:pps}. This is similar to the on-line phase of the system presented in Section~\ref{sec:system}. In particular, step 2 of $\Ppps$ is a modification of the tree-evaluation protocol $\Pte$ where the party with the model knows it in the clear.

\begin{figure}
	\begin{mdframed}
		\begin{itemize}[leftmargin=*,label=--]
			\item \emph{Parties}:  $\cli$ with  $\vec{x}\in\R^n$ and $\MP$ with a  forest $RF=\{T_1,\dots,T_{m}\}$.
			\item \emph{Output}: $y=\frac{1}{m}\sum_{j=1}^{m}T_j(\vec{x})$ for $\cli$.\\
		\end{itemize}
		(\emph{Key-generation}) Given a LHE scheme $(\gen, \enc, \dec)$,  $\cli$ runs $\gen(\kappa)$ and gets $(\sk,\pk)$. The public key $\pk$ is sent to $\MP$. 
		\begin{enumerate}			
			\item \emph{(Encrypted input submission)} $\cli$ computes and sends to $\MP$ the values 
			$\vec{x}'[i]=\enc_\pk(\vec{x}[i])$
			for all $i=1,\dots,n$.
			
			\item \emph{(Tree evaluation)} For each $T_j$, repeat the following. Let $d$ be the depth of $T_j$
			
			\begin{enumerate}
				
				\item \emph{(Randomization)} $\MP$  completes the tree and samples $\gamma_i\leftarrow\{-1,+1\}$, $i=1,\dots,2^d-1$. It proceeds to the tree randomization. From now on, the randomized tree is used.

				\item \emph{($N_i(\vec{x})$ computation} \&  \emph{evaluation)}  For each splitting node $N_i=(j_i,t_i)$, $\MP$ computes
				$N'_i=\vec{x}'[j_i]\odot  t_{i}$. At this point $N_i'=\enc_\pk(N_{i}(\vec{x}))$. $\MP$ and the client  run protocol $\Psc$ where the $\MP$ is Party 1 with input $N'_i$ and the client is Party 2. $\MP$ gets $\alpha_i$ and the client gets $\beta_i$ ($i=1,\dots,2^d-1$).

				\item \emph{(Path computation)} Define $\tilde{\gamma}_i=\gamma_i\alpha_i$, $\MP$ sends $\tilde{\gamma}=(\tilde{\gamma}_1,\dots,\tilde{\gamma}_{2^d-1})$ to the client, and the latter computes 
				$z_i=P_{d,i}^{\tilde{\gamma}}((\beta_j)_{j\in\I_i})$ for all $i=1,\dots,2^d$ (leaf index). Let $i^*$ be the leaf index such that $z_{i^*}\neq0$.
				
				\item \emph{(Label computation)}  $\MP$ samples  $r_{j}\leftarrow\M$ and computes $\tilde{\ell}_i=\ell_i-r_{j}$ for all leaf index $i$. Now $\MP$ and $\cli$ run a $\binom{2^d}{1}$-OT where $\MP$ is the sender with $\tilde{\ell}_1,\dots,\tilde{\ell}_{2^d}$ as input and $\cli$ is the receiver with input $i^*$.  Let $s_j$ be the value received by the client.
				
			\end{enumerate}
			\item \emph{(Prediction)} $\MP$ computes $r= \sum_{j=1}^{m} r_j$ and sends it to the client. The latter computes $y=\frac{1}{m}(\sum_{j=1}^{m} s_j+r)$.
		\end{enumerate}
	\end{mdframed}
	\caption{Protocol $\Ppps$ in the standard privacy-preserving scoring scenario.}
	\label{fig:pps}
\end{figure}

\subsubsection{Complexity and Performance.}
The complexity of protocol $\Ppps$ for a forest with $m$ trees of maximum depth $d$ classifying instances with $n$ features is of $\Theta(n+m\,2^d)$ cryptographic operations. Moreover, the two parties need to communicate $n+2^{d+1}$ ciphertexts between them.
We implement protocol $\Ppps$ using the same setup described in Section~\ref{ssec:efficency};  we test the implementation on the same commodity machine using decision trees trained on six different UCI datasets. 
Table~\ref{tab:tree2} reports the results and shows that the protocol $\Ppps$ has good performance in general. Even for the housing dataset which generates a tree of depth 13, the protocol requires less than 1.2 seconds and 4.3 MB. This is comparable with the results presented in \cite{TMZC17}. There, the running time for the same dataset is  2 seconds and the bandwidth is 0.85 MB. For the nursery dataset, which requires a less deep tree ($d=4$), the running time (resp.\ the bandwidth) we report is roughly 1/5  (resp.\ 1/10)  of the value reported by \cite{TMZC17}.

%

\begin{table}   
	\centering
	\setlength{\tabcolsep}{4pt}
	\renewcommand{\arraystretch}{1.1}
	\begin{tabular}{|l|c|c|c|c|}
		\hline
		Dataset       &  $d$ & $n$ & Time (ms) &  Communication (kB) \\\hline
	Nursery               & 4         & 8             & 29.56    &   10.74       \\
		Heart Disease   & 3         & 13            & 22.77    &  7.75        \\
		Diabetes                & 8         & 9             & 95.25    &  134.66        \\
		Breast cancer          & 4         & 30            & 32.69    &  16.38        \\
		pambase                & 6         & 57            & 63.78    &  48.25        \\
		Housing                & 13        & 13            & 1194.01    &  4200.9 \\\hline       
	\end{tabular}
	\caption{Running time and communication overheard of protocol $\Ppps$ ($m=1$).}
	\label{tab:tree2}
\end{table}


\begin{table}
		\centering
		\setlength{\tabcolsep}{4pt}
		\renewcommand{\arraystretch}{1.3}
		\begin{tabular}{|l|c|c|c|}
			\hline
			& \multicolumn{2}{c|}{Computation} & Communication\\\cline{2-3}
			&  User & Server  & \\ \hline
			Step 1 &  $2^d-1$ encryptions & $n (2^d-1)$ $\mult$ & $2^d-1$ ciphertexts\\ 
			&  $2^d-1$ calls to PRF & $(n+1)(2^d-1)$ $\odot$  & \\ \hline
			Step 3 &   $2^d-1$ decryptions & $2^d-1$ $\mult$  & $2^d-1$ ciphertexts\\ 
			&   					& $2^d-1$ $\boxplus$  & \\\hline
			Step 5 &  $\binom{2^d}{1}$-OT receiver & $\binom{2^d}{1}$-OT sender & $\binom{2^d}{1}$-OT communication\\ 
			&		$1$ decryption	 &  $2^d$ $\boxplus$ & \\ \hline
		\end{tabular}
\caption{Complexity overview  of  $\Pte$ run on a tree of depth $d$ and an input with $n$ features. We use $\boxplus$ to indicate the sum between an encrypted value and a known one.}
\label{tab:complexity} 
\end{table}

\end{document}